\documentclass[draftcls,onecolumn,11pt]{IEEEtran}
\usepackage{subfig}
\usepackage{amssymb}
\usepackage{cite}
\usepackage[pdftex]{graphicx}
\graphicspath{{./}{figures/}}
\usepackage[table]{xcolor}
\usepackage{arydshln}
\usepackage{amsmath}
\usepackage{amssymb}
\usepackage{url}
\usepackage{setspace}
\usepackage{fixltx2e}
\usepackage{rotating}
\usepackage{multirow}
\usepackage{floatrow}
\usepackage[font=small,labelfont=bf]{caption}
\newfloatcommand{capbtabbox}{table}[][\FBwidth]

\newcolumntype{M}{>{$\vcenter\bgroup\hbox\bgroup}c<{\egroup\egroup$}}

\usepackage{blindtext}

\def\ie{\emph{i.e. }}
\def\etal{\emph{et al. }}

\def\mean{\textrm{E}}
\def\var{\textrm{var}}
\def\cov{\textrm{cov}}

\newcommand{\im}[1]{\mathbf{#1}}
\newcommand{\est}[1]{\widehat{#1}}

\def\Pr{\textrm{Pr}}
\usepackage{algorithm}
\usepackage{algorithmic}
\usepackage{caption}
\begin{document}
\setlength{\intextsep}{4pt} 
\setlength{\textfloatsep}{4pt}
\setlength{\abovecaptionskip}{4pt}
\setlength{\belowcaptionskip}{4pt}
\setlength{\dbltextfloatsep}{4pt}
\title{Probabilistic Non-Local Means}
\author{Yue~Wu, {Brian~Tracey}, {Premkumar Natarajan}
        and Joseph~P.~Noonan
\thanks{ Yue Wu, Brian~Tracey and Joseph P. Noonan are with the department of electrical and computer engineering, Tufts university, 161 College Ave, Medford, MA 02155; e-mail: ywu03@ece.tufts.edu. Premkumar Natarajan is with the Raytheon BBN technologies, 10 Moulton St., Cambridge, MA 02138. This project is supported by the Tufts subcontract to BBN on MADCAT contract (Prime Contract HR0011-08-C-004).}}
\maketitle

\begin{abstract}
In this paper, we propose a so-called probabilistic non-local means (PNLM) method for image denoising. Our main contributions are: 1) we point out defects of the weight function used in the classic NLM; 2) we successfully derive all theoretical statistics of patch-wise differences for Gaussian noise; and 3) we employ this prior information and formulate the probabilistic weights truly reflecting the similarity between two noisy patches. The probabilistic nature of the new weight function also provides a theoretical basis to choose thresholds rejecting dissimilar patches for fast computations. Our simulation results indicate the PNLM outperforms the classic NLM and many NLM recent variants in terms of peak signal noise ratio (PSNR) and structural similarity (SSIM) index. Encouraging improvements are also found when we replace the NLM weights with the probabilistic weights in tested NLM variants.
\end{abstract}

\IEEEpeerreviewmaketitle

\section{Introduction}
Non-local means (NLM) is a popular data-adaptive image denoising technique introduced by Buades \etal \cite{NLM0,NLM1}. This technique is proven to be effective in many image denoising tasks. In the classic NLM, a 2D clean image $\im{x}\!=\!\{\!x_l\!\}_{l\in\mathbb I}$ defined on the spatial domain $\mathbb{I}$ is assumed to be contaminated by i.i.d. zero-mean Gaussian noise with an unknown variance $\sigma^2$, \ie
\begin{equation}\label{eq.gaussiannoise}
    y_l = x_l+n_l, \textrm{\, and \,} n_l\sim{\cal{N}}(0,\sigma^2).
\end{equation}
where $y_l$, $x_l$ and $n_l$ denote the noisy observation, the clean image pixel and the pixel noise, respectively.
The NLM then estimates the clean pixel $x_l$ by using a weighted sum of the noisy pixels within a prescribed search region $\mathbb{S}$, typically a square or a rectangular region:
\begin{equation}\label{eq.nlm}
\est{x_l} = \textstyle\sum_{k\in\mathbb{S}_l}{w_{l,k}y_{k}}/W_l
\end{equation}
where each weight is computed by quantifying the similarity between two local patches (defined as $\mathbb{P}$) around noisy pixels $y_l$ and $y_k$ as shown in Eq. \eqref{eq.nlmweight},
\begin{equation}\label{eq.nlmweight}
    w_{l,k} = \exp\big(-{\textstyle\sum_{j\in\mathbb{P}}({y_{l+j}-y_{k+j})^2}/h}\big)
\end{equation}
and the summation of all weights is denoted as
\begin{equation}\label{eq.Wl}
    W_l = \textstyle{\sum_{k\in\mathbb{S}_l}w_{l,k}}.
\end{equation}
Although the original NLM weight \cite{NLM0,NLM1} includes a weak Gaussian smoother, the weight \eqref{eq.nlmweight} is a simplified version with similar performance that is also widely accepted in the NLM community \cite{SURE}.

Within the NLM framework, much progress has been made in recent years. Some authors have focused on fast NLM implementation \cite{vignesh2010fast,6324435}, while others have explored NLM parameter optimization \cite{SURE}, or have adjusted the NLM framework to achieve better performance \cite{NLEM}. We notice that one shared interest of  these three topics is the weight function of the NLM, which is the core of the NLM algorithm.  Calculation of NLM weights is the most computationally expensive part of the algorithm and is related to many parameter optimization schemes. It has long been noticed that the NLM weight function is somewhat inadequate \cite{duval2011bias} because it tends to give non-zero weights to dissimilar patches. However, the reason behind this inadequacy has not fully explored.

In this letter, we focus on the NLM weight function and propose a new weight under a probabilistic framework. The rest of the paper is organized as follows: Sec. II shows the defects of the NLM weights; Sec. III proposes our PNLM framework with new probabilistic weights; Sec. IV shows simulation results; and we conclude the letter in Sec. V.
\section{Problems with the NLM Weight Function}
The NLM weight function \eqref{eq.nlmweight} is considered as $w_{l,k}\!=\!\exp(-\!D_{l,k}\!/h')$, where $h'$ is a translation of the temperature parameter $h$ in \eqref{eq.nlmweight} and
\begin{equation}\label{eq.bigD0}
    D_{l,k} = \textstyle\sum_{j\in\mathbb{P}}(y_{l+j}-y_{k+j})^2/2\sigma^2
\end{equation}
is the patch difference between the patches around $y_l$ and $y_k$. In this way, \eqref{eq.bigD0} can be interpreted as the standard quantitative $\chi^2$ test to measure the similarity of the two samples \cite{statisticalNLM}. The statistical interpretation of the exponential function used in \eqref{eq.nlmweight} is not straightforward \cite{statisticalNLM}, although may be possible to relate it to Gaussian kernels used in probability density estimation. Nevertheless, this exponential function gives a larger weight to a pixel with a smaller patch difference (Fig. 1(a)). Intuitively, this idea is quite reasonable, as it means that the NLM relies more on pixels with smaller patch differences. However, we demonstrate below that this exponential function makes the NLM weights somewhat problematic.

\begin{figure}[h]
  \includegraphics[width=.618\linewidth]{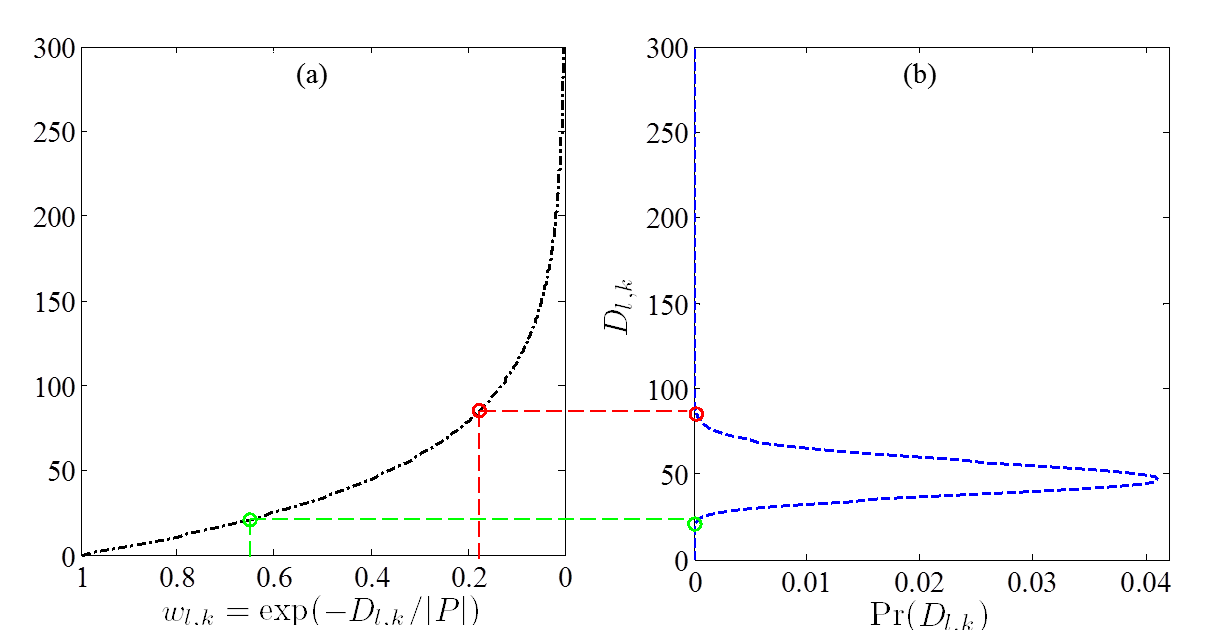}\\
  \caption{NLM weight function and the distribution of 7$\times$7 patch differences. (a) the NLM weight for $h=\sigma^2|\mathbb{P}|$; and (b) the distribution of disjoint patch differences. Red and green circles denotes two equal probable patch differences, while are biasedly weighted in NLM. }\label{fig.NLMweight}
\end{figure}

From now on, we consider $D_{l,k}$ as a random variable (r.v.) and assume patches around $x_l$ and $x_k$ match perfectly, \ie
 \begin{equation}\label{eq.match}
\textstyle\sum_{j\in\mathbb{P}}(x_{l+j}\!-\!x_{k+j})^2\!=\!0.
\end{equation}
If they are disjoint, then $D_{l,k}\!\!\sim\!\!\chi^2_{|\mathbb{P}|}$, where $|\!\cdot\!|$ denotes the cardinality function. Fig. 1 shows this distribution on the right, with the corresponding NLM weight function on the left. It is clear that the NLM weight function gives two equally probable $D_{l,k}$s very different weights and that it fails to give the largest weight to the most probable case. For $D_{l,k}$s close to its expected value, the weight errors are not too large because the corresponding region in the exponential curve is almost linear with a moderate slope. However, for $D_{l,k}$s far away from its expected value, weight errors are very large because the NLM function tends to give nonzero weights to these highly improbable cases, so weight errors grow quickly. In practice, correcting over-weighted weights has been shown to improve NLM performance. For example, the center pixel weight (CPW) in NLM is unitary and thus over-weights center pixels. \cite{CPWstein,JSCPW} report noticeable improvement just by tuning these over-weighted CPWs.

\section{Probabilistic Non-Local Means }
Instead of including the exponential function in weighting pixels, we propose the following probabilistic weight
\begin{equation}\label{eq.nlmweight3}
    w_{l,k} = f_{l,k}\big(\est{D}_{l,k}/\rho^2\big)
\end{equation}
where $f_{l,k}(\cdot)$ is the theoretical probability density function (p.d.f.) of the r.v. $D_{l,k}$, $\est{D}_{l,k}$ is the patch difference using estimated variance $\est{\sigma}^2$ in \eqref{eq.bigD0}, and $\rho$ is a tuning parameter. This weight function then can be interpreted as the probability of seeing a noisy patch difference when we know the two clean patches match perfectly.

 \vspace{-5pt}
\subsection{Theoretical Distribution of Patch-wise Distance}
Pretend we know the true noise variance $\sigma^2$ (later we will show this knowledge is unnecessary). Our goal is to derive the theoretical p.d.f. of the patch difference when the two clean patches around pixel $x_l$ and $x_k$ are perfectly matching (see \eqref{eq.match}). To do so, we denote the pixel distance $d_{l,k}$ as
\begin{equation}\label{eq.dlk}
    d_{l,k}\!=\!(y_l-y_k)^2/2\sigma^2
\end{equation}
and thus we have $D_{l,k}$ of the form that
\begin{equation}\label{eq.bigD}
    D_{l,k} = \textstyle\sum_{j\in\mathbb{P}}d_{l+j,k+j}.
\end{equation}
Because two patches are perfectly matching and noise is i.i.d., for all $j\in\mathbb{P}$ we have $$d_{l+j,k+j}={(n_{l+j}-n_{k+j})^2\over 2\sigma^2}\sim\chi_1^2.$$
If all $d_{l+j,k+j}$s are i.i.d., then we have $
    D_{l,k} {\sim}{\chi^2_{|\mathbb{P}|}}$,
whose mean is $|\mathbb{P}|$ and variance is $2|\mathbb{P}|$. However, the i.i.d. assumption does not hold when the two patches overlap, as is the case for many pairs of patches. Fortunately, it is known that such a summed correlated $\chi^2$ distribution can be well approximated \cite{sumChi} as follows,
\begin{equation}\label{eq.Dniid}
    D_{l,k} \sim{\gamma_k\chi^2_{\eta_{k}}}
\end{equation}
where parameters $\gamma_k $ and $\eta_{k}$ can be determined by the first two cumulants of $D_{l,k}$ \cite{sumChi} as shown below.
\begin{equation}\label{eq.gamma}
    \gamma_k = \var[D_{l,k}]/(2\mean[D_{l,k}])
\end{equation}
\begin{equation}\label{eq.eta}
    \eta_{k} = \mean[D_{l,k}]/\gamma_k
\end{equation}The cumulant $\mean[D_{l,k}]$ is straightforward to find, and it is
\begin{equation}\label{eq.Dlkmean}
\mean[D_{l,k}]\!=\textstyle\sum_{j\in\mathbb{P}}\mean[d_{l+j,k+j}]=\!|\mathbb{P}|.
\end{equation}
With regards to $\var[D_{l,k}]$ , the following identity always holds
\begin{equation}\label{eq.Dvar}
    \var[D_{l,k}] =\textstyle \sum_{i,j\in\mathbb{P}}\cov[d_{l+i,k+i},d_{l+j,k+j}]
\end{equation}
where the covariance can be written as follows.
\begin{equation}\label{}
    \cov[d_{l+i,k+i},d_{l+j,k+j}] = \mean[d_{l+i,k+i}d_{l+j,k+j}]-\mu_{d_{l,k}}^2
\end{equation}
This equation compares two pairs of r.v.s,  $\mathbb{N}_{l,k}^i\!\!=\!\!\{n_{l+i},n_{k+i}\}$ and $\mathbb{N}_{l,k}^j\!\!=\!\!\{n_{l+j},n_{k+j}\}$, of which 0, 1 or 2 may be repeated.  Thus,
\begin{equation}\label{eq.meandlilj}
    \mean[d_{l+i,k+i}d_{l+j,k+j}] = \left\{
    \begin{array}{r}
    3, \textrm{if}\, |\mathbb{N}_{l,k}^i\cap\mathbb{N}_{l,k}^j|=2\\
    1.5,\textrm{if}\, |\mathbb{N}_{l,k}^i\cap\mathbb{N}_{l,k}^j|=1\\
    1, \textrm{if}\, |\mathbb{N}_{l,k}^i\cap\mathbb{N}_{l,k}^j|=0
    \end{array}
    \right.
\end{equation}
found as $\mean[d_{l+i,k+i}d_{l+j,k+j}]$ measures kurtosis if both pixels are repeated; pixels are i.i.d. if distinct; and by expanding terms if one pixel is repeated.  This implies that
\begin{equation}\label{eq.meandlilj}
    \cov[d_{l+i,k+i}d_{l+j,k+j}] = \left\{
    \begin{array}{r}
    2, \textrm{if}\, |\mathbb{N}_{l,k}^i\cap\mathbb{N}_{l,k}^j|=2\\
    0.5,\textrm{if}\, |\mathbb{N}_{l,k}^i\cap\mathbb{N}_{l,k}^j|=1\\
    0, \textrm{if}\, |\mathbb{N}_{l,k}^i\cap\mathbb{N}_{l,k}^j|=0
    \end{array}
    \right..
\end{equation}
$\var[D_{l,k}]$ is thus dependent on the number of terms for which $|\mathbb{N}_{l,k}^i\!\!\cap\!\mathbb{N}_{l,k}^j|\!=\!2$ and for which $|\mathbb{N}_{l,k}^i\!\!\cap\!\mathbb{N}_{l,k}^j|\!=\!1$. Case $|\mathbb{N}_{l,k}^i\!\!\cap\!\mathbb{N}_{l,k}^j|\!=\!2$ happens only when $i\!=\!j$, and so the number of overlapped pixels is $|\mathbb{P}|$. Case $|\mathbb{N}_{l,k}^i\!\!\cap\!\!\mathbb{N}_{l,k}^j|\!=\!1$ happens only when two patches overlap, and the corresponding number is the same as the number of overlapping pixels. Letting $\mathbb{O}_{l,k}$ be the set of overlapping pixels, $\var[D_{l,k}]$ can be written as
\begin{equation}\label{eq.varDlke}
    \var[D_{l,k}] = 2|\mathbb{P}|+|\mathbb{O}_{l,k}|
\end{equation}
and is known once $k$ is given. As a result, $\gamma_k$ and $\eta_k$ in \eqref{eq.gamma} and \eqref{eq.eta} can be determined, implying the p.d.f. of $D_{l,k}$ is
\begin{equation}\label{eq.pdfdlk}
    f_{l,k}(D)\!=\!\chi^2_{\eta_k}(D/\gamma_k)\!=\!\dfrac{(D/\gamma_k)^{\eta_k/2-1}\exp(-D/2\gamma_k)}{2^{\eta_k/2}\Gamma(\eta_k/2)}.
\end{equation}
The different spatial relationships of patch pairs imply different $|\mathbb{O}_{l,k}|$, thus causing different $\var[D_{l,k}]$,  $\gamma_k$, and finally  p.d.f. $f_{l,k}$. This conclusion means that NLM weights calculated without considering spatial correlations are inadequate.

$f_{l,k}$ also provides a natural criterion for speeding computation by rejecting over-dissimilar patches at an early stage (see similar usage in \cite{vignesh2010fast}). Thresholds can be set by finding critical values $D^{*+}_\alpha$ and $D^{*-}_{\alpha}$ under a prescribed significance level $\alpha$ such that
\begin{equation}\label{eq.cvalues}
    \Pr(D^{*-}_\alpha
    \leq D\leq D^{*+}_\alpha | f_{l,k}) = \alpha.
\end{equation}

\subsection{Parameters Discussions}
Above we did not use $f_{l,k}(\est{D}_{l,k})$ as our weight function, but instead used $f_{l,k}(\est{D}_{l,k}/\rho^2)$. The parameter $\rho^2$ provides a way to adjust our probabilistic model when an estimated variance $\est{\sigma}^2$ is used instead of the true $\sigma^2$. When \begin{equation}\label{eq.rhosq}
\rho^2 = \sigma^2/\est{\sigma}^2
\end{equation}
reflects the ratio of the true noise variance to the estimated one, all previous derivations  hold. Thus \begin{equation}\label{eq.dlk}
    \est{d}_{l,k}\!=\!(y_l-y_k)^2/2\est{\sigma}^2\!=\!\rho^2(y_l-y_k)^2/2{\sigma}^2\sim\chi^2_{\rho^2}
\end{equation}
indicating that ${\mean[\est{D}_{l,k}]}\!=\!\rho^2|\mathbb{P}|$ and ${\var[\est {D}_{l,k}]}\!=\!\rho^4(2|\mathbb{P}|\!+\!|\mathbb{O}_{l,k}|)$. Further, this implies that the actual parameters used are $\est{\gamma}_k\!=\!\rho^2\gamma_k$ and $\est{\eta}_k\!=\!\eta_k$. Finally, these results lead to\vspace{-5pt}
\begin{equation}\label{}
    \nonumber\est{f}_{l,k}(\est{D}) = \chi^2_{\est{\eta}_k}(\est{D}/\est{\gamma}_k) = \chi^2_{\eta_k}(\est{D}/(\rho^2{\gamma}_k))= f_{l,k}(\est{D}/\rho^2)
\end{equation}
which is the weight function given in \eqref{eq.nlmweight3}.

The raw probabilistic CPW $w_{l,l}\!=\!f_{l,k}(0)\!\approx\!0$ under-weights a noisy center pixel. A more plausible CPW is
\begin{equation}\label{eq.cpw}
    w_{l,l} = \chi^2_{|\mathbb{P}|}(|\mathbb{P}|).
\end{equation}
which is the same as the weight of the most probable case. This CPW is used for the rest of the letter.
\section{Simulation Results}
All of the following simulations are done under the MATLAB r2012b environment. Our two goals are 1) to show that the derived p.d.f. $f_{l,k}$ in \eqref{eq.pdfdlk} closely approximates its true p.d.f.; and 2) to confirm the superiority of the proposed probabilistic weights and PNLM.

Fig. 2 shows the $\var[D_{l,k}]$ map for a 7$\times$7 search region $\mathbb{S}_l$ with 3$\times$3 patches and the six typical theoretical p.d.f.s $f_{l,k}$, plotted with the corresponding sample distributions estimated from 100,000 realizations. It is noticeable that the $\var[D_{l,k}]$ map is location-dependent and isotropic with one of the six theoretical values $\{$18,19,20,21,22,24$\}$. The more pixels overlap, the larger $\var[D_{l,k}]$ is, implying a smaller peak on its p.d.f. It is clear that the predicted p.d.f.s are very close to those estimated from a large number of samples.

Since it is clear that the accuracy of the $f_{l,k}$ approximation degrades as correlation increases, the approximation accuracy of the most-overlapped cases can be used to characterize the worst-case accuracy.
For each combination of search region $\mathbb{S}_l$ and patch size $\mathbb{P}$, there are four possible $k$s that attain the maximum correlation, all of which are one pixel away from the center pixel (see examples for $k$=18, 24, 26, and 32 on Fig. 2-(a)). In Table I, we report the averaged P-values of goodness of fit tests for the most correlated $f_{l,k}$s, where each P-value is the averaged from P-values of the four most correlated $f_{l,k}$s. Because all observed P-values are above 5\%, we say the approximated theoretical p.d.f. \eqref{eq.pdfdlk} gives satisfactory predictions, so these p.d.f.s can reliably be used to quantify patch similarities.

\begin{figure}[h]
  \scriptsize
\includegraphics[width=.4\linewidth]{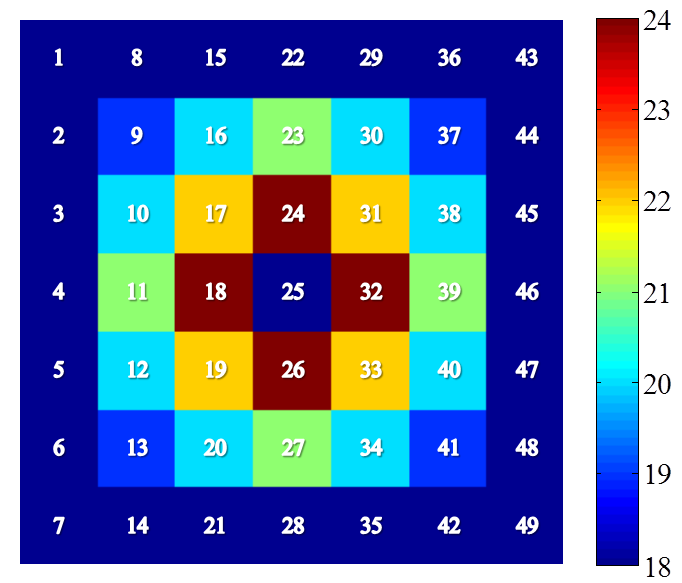}{(a)}\\
  \begin{tabular}{@{}M@{}M@{}M@{}}
\includegraphics[width=4cm]{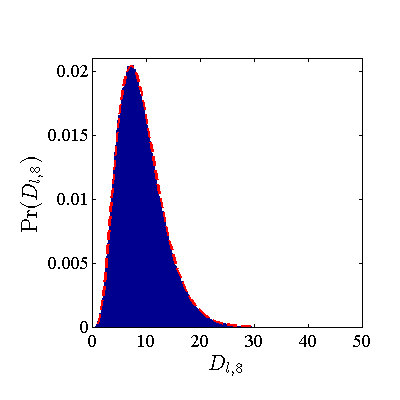}{$\!\!\!\!\!$(b)}& \includegraphics[width=4cm]{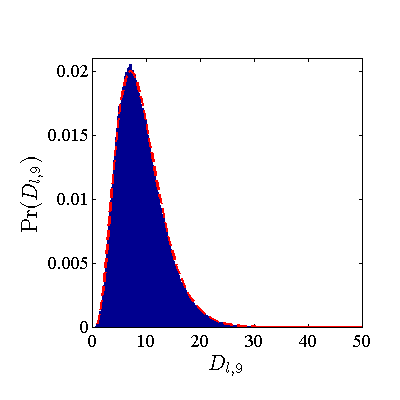}{$\!\!\!\!\!$(c)}& \includegraphics[width=4cm]{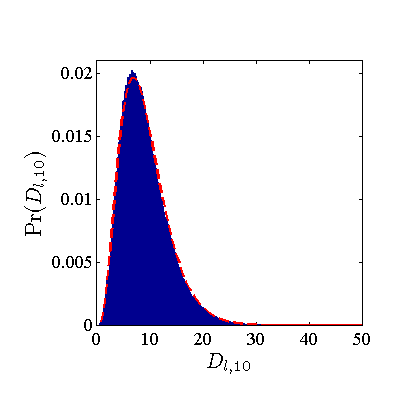}{$\!\!\!\!\!$(d)}
  \end{tabular}
  \begin{tabular}{@{}M@{}M@{}M@{}}
\includegraphics[width=4cm]{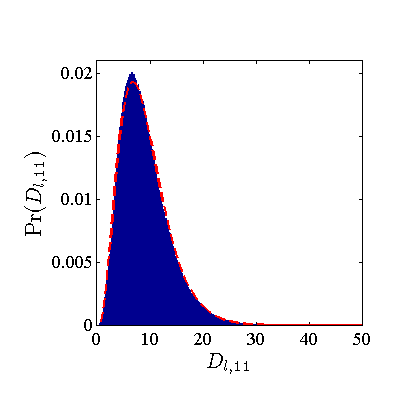}{$\!\!\!\!\!$(e)}& \includegraphics[width=4cm]{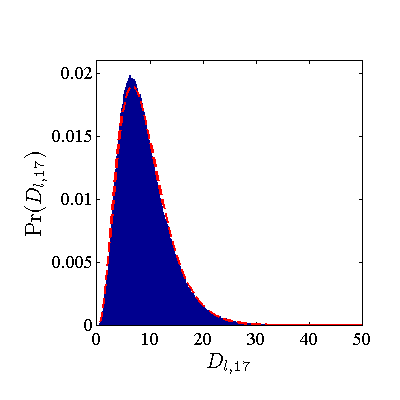}{$\!\!\!\!\!$(f)} & \includegraphics[width=4cm]{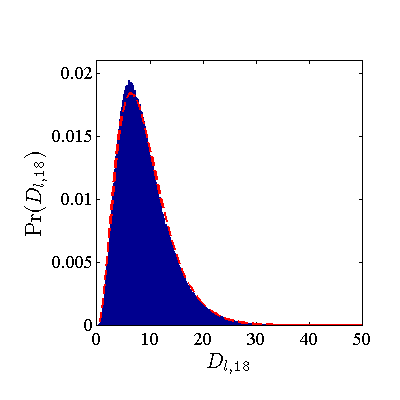}{$\!\!\!\!\!$(g)}
  \end{tabular}
  \caption{Theoretical and estimated p.d.f. $f_{l,k}$s for 7$\times$7 search region and 3$\times$3 patches. (a) theoretical $\var[D_{l,k}]$ map (white indices indicates $k$s in $\mathbb{S}_l$). (b)-(g) theoretical (red dash lines) and estimated (blue bars) p.d.f. $f_{l,k}$s for $k\!=\!8$ ($\var[D_{l,8}]\!\!=\!\!18$), $k\!=\!9$ ($\var[D_{l,9}]\!=\!19$),$k\!=\!10$ ($\var[D_{l,10}]\!=\!20$),$k\!=\!11$ ($\var[D_{l,11}]\!=\!21$), $k\!=\!17$ ($\var[D_{l,17}]\!=\!22$) and $k\!=\!18$ ($\var[D_{l,18}]\!=\!24$), respectively.}
\end{figure}

\begin{table}[h]
\caption{Averaged P-values of goodness of fit tests for the observed sample distributions.}
\centering\scriptsize
\begin{tabular}{r|r|ccccc}
  \hline
  \multicolumn{2}{c|}{}&\multicolumn{5}{c}{\bf Search Region Size}\\\cline{3-7}
   \multicolumn{2}{c|}{}& 7 & 11 & 15 & 21 & 29 \\\hline
  \multirow{4}{*}{\begin{sideways}{\bf Patch Size}\end{sideways}}& 3&0.5853  &  0.6865    &0.2252   & 0.1001 &   0.5612\\
   &5&0.2675   & 0.3125    &0.3374   & 0.5746 &   0.3501\\
   &7 &0.3967   & 0.4659    &0.1545   & 0.2645 &   0.4282\\
   &9 &   0.4741   & 0.8665   & 0.5233   & 0.3405  &  0.6582\\
  \hline
\end{tabular}
\end{table}

In the following simulation, we compare the three pairs of NLM and PNLM algorithms, namely 1) the classic NLM and the proposed PNLM, 2) the classic NLM with the James-Stein Shrinkage (JSNLM) \cite{JSCPW} and the proposed PNLM with the James-Stein Shrinkage (PSJNLM), and 3)the nonlocal median with the classic weights (NLEM) \cite{NLEM} and the nonlocal median with the probabilistic weights (PNLEM). The only difference between the two algorithm in each pair is the weight function. With regards to the parameter settings, we use patch size 7 and search region size 21 for all methods. For the temperature parameter $h$ in NLMs, we use $h\!=\!|\mathbb{P}|\sigma^2$, which is nearly optimal and suggested in \cite{SURE}. For $\rho$ in PNLMs, we use $\rho\!=\!1$. To quantify the quality of a denoised method, we compute the average PSNR \cite{SURE} and SSIM \cite{ssim} scores from 10 realizations for each method and each noise level. These results are reported in Table II.

\begin{table}[h]
\caption{Performance comparisons for NLM and PNLM methods}
\scriptsize\centering
\begin{tabular}{@{}M@{}m{.1cm}@{}M@{}m{.1cm}@{}|@{}m{.1cm}@{}M@{}m{.1cm}@{}M@{}m{.1cm}@{}M@{}m{.1cm}@{}M@{}m{.1cm}@{}M@{}m{.1cm}@{}M@{}m{.1cm}@{}M@{}m{.1cm}@{}M@{}m{.1cm}@{}M@{}m{.1cm}@{}M@{}}
  \hline\hline
  \multicolumn{3}{r}{\bf PSNR(dB)$\backslash$ $\sigma$}&&&10&&20&&30&&40&&50&&60&&70&&80&&90&&100\\\hline
  \multirow{6}{*}{\begin{sideways}{\bf\textit{cameraman}}\end{sideways}} &&NLM&&&32.57&&28.92&&26.98&&24.98	 &&23.52&&22.52&&21.84&&21.24&&20.82&&20.44\\
  &&PNLM&&&32.47&&29.08&&\bf{27.44}&&\bf{26.26}&&\bf{25.19}&&\bf{24.13}&&\bf{23.26}&&\bf{22.44}&&\bf{21.84}&&\bf{21.31}\\
  &&NLEM&&&32.66&&28.90&&26.63&&24.78&&23.35&&22.16&&21.78&&21.31&&20.95&&20.55\\
  &&PNLEM&&&\bf{33.06}&&\bf{29.42}&&27.36&&25.72&&24.90&&23.87&&23.11&&22.28&&21.73&&21.13\\
  &&JSNLM&&&32.64&&29.01&&27.13&&25.46&&24.12&&23.10&&22.33&&21.61&&21.09&&20.63\\
  &&PSJNLM&&&32.21&&29.08&&27.45&&26.16&&25.07&&24.05&&23.20&&22.40&&21.77&&21.23\\\hline
  \multirow{6}{*}{\begin{sideways}{\bf\textit{house}}\end{sideways}}
  &&NLM&&&34.08&&31.30&&28.79&&26.88&&25.62&&24.66&&23.85&&23.31&&22.90&&22.45\\
  &&PNLM&&&\bf{34.92}&&\bf{32.40}&&\bf{30.48}&&\bf{28.70}&&\bf{27.25}&&\bf{26.14}&&\bf{24.98}&&\bf{24.17}&&\bf{23.57}&&\bf{22.98}\\
  &&NLEM&&&34.30&&30.43&&27.80&&26.53&&25.51&&24.88&&24.13&&23.52&&22.92&&22.43\\
  &&PNLEM&&&34.56&&31.97&&30.24&&28.64&&27.07&&{26.11}&&24.90&&24.14&&23.38&&22.94\\
  &&JSNLM&&&34.62&&31.70&&29.29&&27.30&&25.94&&24.87&&23.98&&23.35&&22.86&&22.37\\
  &&PSJNLM&&&34.81&&32.38&&30.36&&28.58&&27.11&&25.89&&24.82&&23.99&&23.36&&22.75\\
\hline  \multirow{6}{*}{\begin{sideways}{\bf\textit{lenna}}\end{sideways}}
  &&NLM&&&33.74&&30.91&&28.72&&27.14&&26.03&&25.13&&24.42&&23.88&&23.44&&23.03\\
  &&PNLM&&&34.59&&\bf{32.07}&&\bf{30.17}&&\bf{28.58}&&\bf{27.32}&&\bf{26.23}&&\bf{25.33}&&\bf{24.59}&&\bf{23.98}&&\bf{23.43}\\
  &&NLEM&&&33.56&&30.00&&28.41&&27.30&&26.47&&25.65&&25.05&&24.29&&23.64&&23.12\\
  &&PNLEM&&&33.78&&31.21&&29.64&&28.32&&27.26&&26.28&&25.58&&24.80&&24.20&&23.69\\
  &&JSNLM&&&34.38&&31.41&&29.21&&27.49&&26.26&&25.26&&24.47&&23.85&&23.33&&22.86\\
  &&PSJNLM&&&\bf{34.72}&&\bf{32.07}&&30.09&&28.48&&27.20&&26.07&&25.15&&24.38&&23.74&&23.15\\\hline  \multirow{6}{*}{\begin{sideways}{\bf\textit{checker}}\end{sideways}}
  &&NLM&&&39.04&&33.80&&30.95&&28.94&&27.37&&25.94&&24.45&&23.25&&21.84&&20.87\\
  &&PNLM&&&\bf{40.34}&&\bf{35.17}&&\bf{32.31}&&\bf{30.26}&&\bf{28.38}&&\bf{26.71}&&\bf{25.37}&&\bf{24.50}&&\bf{23.29}&&\bf{22.76}\\
  &&NLEM&&&39.68&&34.13&&30.90&&28.94&&27.07&&25.62&&24.59&&23.54&&22.66&&22.09\\
  &&PNLEM&&&39.72&&34.49&&31.25&&29.18&&27.15&&25.77&&24.68&&23.78&&22.93&&22.52\\
  &&JSNLM&&&39.03&&33.79&&30.93&&28.93&&27.35&&25.91&&24.42&&23.22&&21.81&&20.84\\
  &&PSJNLM&&&34.64&&30.86&&31.23&&29.73&&27.95&&26.39&&25.08&&24.25&&23.10&&22.60\\\hline\hline
  \multicolumn{3}{r}{\bf SSIM(\%)$\backslash$ $\sigma$}&&&10&&20&&30&&40&&50&&60&&70&&80&&90&&100\\\hline
  \multirow{6}{*}{\begin{sideways}{\bf\textit{cameraman}}\end{sideways}}
  &&NLM&&&91.08&&82.92&&78.50&&73.87&&68.97&&64.18&&59.78&&55.58&&51.87&&48.69\\
  &&PNLM&&&\bf{91.64}&&\bf{84.65}&&\bf{80.23}&&\bf{76.61}&&\bf{73.26}&&\bf{69.72}&&\bf{66.18}&&\bf{62.72}&&\bf{59.45}&&\bf{56.60}\\
  &&NLEM&&&88.68&&80.24&&72.53&&64.98&&59.15&&53.04&&48.67&&43.96&&40.45&&35.95\\
  &&PNLEM&&&91.16&&83.22&&78.13&&73.31&&68.89&&63.11&&58.90&&54.51&&50.54&&46.18\\
  &&JSNLM&&&91.23&&84.32&&78.91&&73.63&&68.74&&64.01&&59.62&&55.34&&51.36&&48.23\\
  &&PSJNLM&&&89.69&&84.04&&79.29&&74.97&&71.01&&66.98&&63.13&&59.30&&55.61&&52.82\\\hline  \multirow{6}{*}{\begin{sideways}{\bf\textit{house}}\end{sideways}}
  &&NLM&&&87.63&&83.77&&79.88&&75.11&&70.63&&66.34&&62.16&&58.30&&54.86&&51.40\\
  &&PNLM&&&\bf{89.38}&&\bf{85.00}&&\bf{81.72}&&\bf{78.18}&&\bf{74.70}&&\bf{71.05}&&\bf{67.43}&&\bf{63.99}&&\bf{60.81}&&\bf{58.05}\\
  &&NLEM&&&88.06&&81.79&&75.43&&69.11&&63.09&&57.09&&51.09&&45.81&&41.19&&37.04\\
  &&PNLEM&&&89.17&&84.11&&79.92&&75.17&&70.40&&65.40&&59.92&&54.59&&49.58&&46.80\\
  &&JSNLM&&&89.12&&84.14&&79.54&&74.52&&69.78&&65.19&&60.80&&56.94&&53.33&&49.86\\
  &&PSJNLM&&&89.34&&84.57&&80.64&&76.45&&72.35&&68.03&&63.80&&60.11&&56.49&&53.34\\\hline
    \multirow{6}{*}{\begin{sideways}{\bf\textit{lenna}}\end{sideways}}
    &&NLM&&&87.86&&83.98&&79.39&&74.90&&70.72&&66.72&&62.87&&59.23&&55.82&&52.59\\
  &&PNLM&&&89.69&&\bf{85.00}&&\bf{81.18}&&\bf{77.56}&&\bf{74.16}&&\bf{70.78}&&\bf{67.57}&&\bf{64.51}&&\bf{61.44}&&\bf{58.81}\\
  &&NLEM&&&88.21&&81.19&&75.44&&69.48&&63.71&&57.57&&52.22&&46.92&&42.29&&38.54\\
  &&PNLEM&&&89.56&&84.02&&79.03&&74.05&&69.42&&64.64&&60.09&&55.85&&51.87&&48.25\\
  &&JSNLM&&&89.37&&83.94&&79.16&&74.48&&69.95&&65.67&&61.61&&57.86&&54.32&&50.99\\
  &&PSJNLM&&&\bf{89.75}&&84.64&&80.21&&76.00&&71.93&&67.97&&64.16&&60.70&&57.23&&54.11\\\hline
    \multirow{6}{*}{\begin{sideways}{\bf\textit{checker}}\end{sideways}}
    &&NLM&&&99.01&&97.38&&95.35&&93.23&&90.75&&87.94&&84.22&&80.76&&76.31&&72.00\\
  &&PNLM&&&\bf{99.33}&&\bf{98.32}&&\bf{97.03}&&\bf{95.68}&&\bf{93.82}&&\bf{91.79}&&\bf{89.48}&&\bf{87.61}&&\bf{84.87}&&\bf{82.95}\\
  &&NLEM&&&99.10&&97.67&&95.46&&92.65&&89.20&&85.12&&81.90&&78.73&&74.26&&71.71\\
  &&PNLEM&&&99.22&&97.99&&96.03&&94.26&&91.83&&88.85&&85.91&&83.16&&79.38&&77.35\\
  &&JSNLM&&&99.00&&97.34&&95.24&&93.11&&90.53&&87.59&&83.79&&80.23&&75.65&&71.18\\
  &&PSJNLM&&&94.25&&92.72&&95.09&&93.74&&91.66&&89.53&&86.86&&84.83&&82.02&&80.14\\\hline
  \hline
\end{tabular}
\end{table}

From Table II, it is clear that 1) the proposed PNLM method outperforms the NLM method and those recent variants like NLEM and JSNLM; and 2) by replacing the NLM weight with the new proposed probabilistic one, both NLEM and JSNLM are improved in terms of higher PSNR/SSIM scores. Fig.3 gives example denoising results and method noise images of the NLM and PNLM algorithms. These results show that the effectiveness of the new proposed probabilistic weight and the superiority of the PNLM framework.

\begin{figure}[!h]
  \scriptsize\centering
  \begin{tabular}{@{}r@{}M@{}m{.05cm}@{}M@{}r@{}M@{}m{.05cm}@{}M}

(a)&\includegraphics[width=2cm]{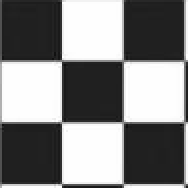} && & (b)&\includegraphics[width=2cm]{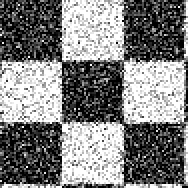} &&\\
(c)&\includegraphics[width=2cm]{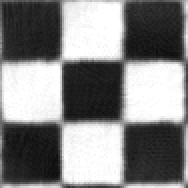} && \includegraphics[width=2cm]{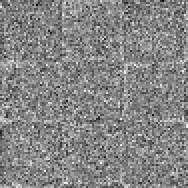} & (d)&\includegraphics[width=2cm]{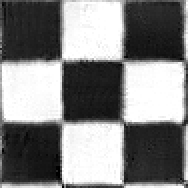} &&\includegraphics[width=2cm]{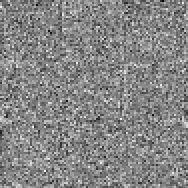}\\
(e)&\includegraphics[width=2cm]{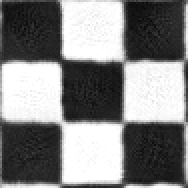} && \includegraphics[width=2cm]{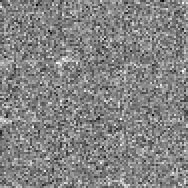} & (f)&\includegraphics[width=2cm]{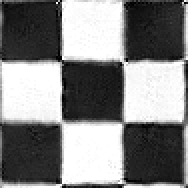} &&\includegraphics[width=2cm]{rpnlm}\\
(g)&\includegraphics[width=2cm]{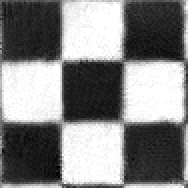} && \includegraphics[width=2cm]{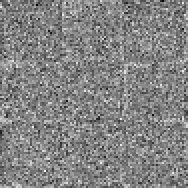} & (h)&\includegraphics[width=2cm]{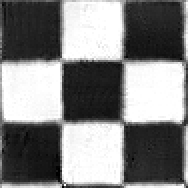} &&\includegraphics[width=2cm]{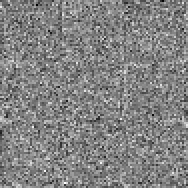}\\
  \end{tabular}
  \caption{NLM and PNLM denoising results for $\sigma\!=\!80$ (cropped and enlarged from results of image \textit{checker}). (a) clean image; (b) noisy observation; (c) to (h): denoising results and method noise images of NLM, PNLM, NLEM, PNLEM, JSNLM, and PJSNLM, respectively.}
\end{figure}

\section{Conclusion}
In this letter, we pointed out the insufficiency of the NLM weights and showed a new promising PNLM framework, whose weights better reflect patch similarities. The proposed PNLM framework connects the denoising process and the noise type and thus is meaningful for denoising other types of noise. As long as a noise p.d.f. is known, we can estimate $f_{l,k}$ correspondingly. In this way, a universal denoising framework (see example \cite{sun2012modifying}) for multiple types of known noises and mixed noises may be developed. In addition, the proposed  PNLM can also be extended to capture non i.i.d. noises, because one can easily to replace the p.d.f. of patch difference $f_{l,k}$ with more general forms. For example, for Gaussian noises with changing variance, $f_{l,k}(D|\sigma^2)\Pr(\sigma^2)$ can be used in place of $f_{l,k}(D)$. The proposed PNLM also provides a theoretical basis to quantify patch similarities, so Eq. \eqref{eq.cvalues} can be used in other ways other than early termination. For example, the critical values predicted by $f_{l,k}$ can also be used as thresholds to reject or accept a patch in the first stage of BM3D \cite{dabov2006image}. This choice provides a theoretically-based alternative to the empirical hard thresholds in BM3D (\ie $\tau_{match}$ in Eq. (2) of \cite{dabov2006image}). In our initial tests, we also see a performance improvement after using the probabilistic thresholds.
\bibliographystyle{IEEEtran}
\bibliography{report}
\end{document}